\documentclass[final,authoryear,5p,times,twocolumn]{elsarticle}
\pdfoutput=1

\usepackage{graphicx}
\usepackage{subfig}
\usepackage[cmex10]{amsmath}
\usepackage{amssymb}
\usepackage{multirow}
\usepackage{url}
\usepackage{placeins}
\usepackage{floatrow}
\usepackage[dvipsnames]{xcolor}
\usepackage{color}
\usepackage{flushend}

\usepackage{tabularx}
\usepackage{array}
\usepackage{rotating}

\usepackage{arydshln}
\usepackage{enumitem}

\newcommand{\added}[1]{\textcolor{black}{#1}}
\newcommand{\changed}[1]{\textcolor{black}{#1}}

\newcommand{\explain}{ExplAIn}

\newcolumntype{Y}{>{\centering\arraybackslash}X}
\newcommand\RotText[1]{\rotatebox{90}{\parbox{2.2cm}{\centering#1}}}

\journal{Medical Image Analysis}

\begin{document}

\begin{frontmatter}



\title{\explain: Explanatory Artificial Intelligence for Diabetic Retinopathy Diagnosis}

\author[label1]{Gwenol\'e~Quellec\corref{cor1}}
\ead{gwenole.quellec@inserm.fr}
\author[label2,label1]{Hassan~Al~Hajj}
\author[label2,label1]{Mathieu~Lamard}
\author[label3,label1]{Pierre-Henri~Conze}
\author[label4]{Pascale~Massin}
\author[label2,label1,label5]{B\'eatrice~Cochener}
\cortext[cor1]{LaTIM - IBRBS - 22, avenue Camille Desmoulins - 29200 Brest, France - Tel.: +33 2 98 01 81 29}
\address[label1]{Inserm, UMR 1101, Brest, F-29200 France}
\address[label2]{Univ Bretagne Occidentale, Brest, F-29200 France}
\address[label3]{IMT Atlantique, Brest, F-29200 France}
\address[label4]{Service d'Ophtalmologie, H\^{o}pital Lariboisi\`ere, APHP, Paris, F-75475 France}
\address[label5]{Service d'Ophtalmologie, CHRU Brest, Brest, F-29200 France}

\begin{abstract}
  In recent years, Artificial Intelligence (AI) has proven its relevance for medical decision support. However, the ``black-box'' nature of successful AI algorithms still holds back their wide-spread deployment. In this paper, we describe an eXplanatory Artificial Intelligence (XAI) that reaches the same level of performance as black-box AI, for the task of classifying Diabetic Retinopathy (DR) severity using Color Fundus Photography (CFP). This algorithm, called \explain, learns to segment and categorize lesions in images; the final image-level classification directly derives from these multivariate lesion segmentations. The novelty of this explanatory framework is that it is trained from end to end, with image supervision only, just like black-box AI algorithms: the concepts of lesions and lesion categories emerge by themselves. For improved lesion localization, foreground/background separation is trained through self-supervision, in such a way that occluding foreground pixels transforms the input image into a healthy-looking image. The advantage of such an architecture is that automatic diagnoses can be explained simply by an image and/or a few sentences. \explain~is evaluated at the image level and at the pixel level on various CFP image datasets. We expect this new framework, which jointly offers high classification performance and explainability, to facilitate AI deployment.
\end{abstract}


\end{frontmatter}

\section{Introduction}
\label{sec:Introduction}

Diabetic Retinopathy (DR) is a leading and growing cause of vision impairment and blindness: by 2040, around 600 million people throughout the world will have diabetes \citep{ogurtsova_idf_2017}, a third of whom will have DR \citep{yau_global_2012}. Early diagnosis is key to slowing down the progression of DR and therefore preventing the occurrence of blindness. Annual retinal screening, generally using Color Fundus Photography (CFP), is thus recommended for all diabetic patients \citep{javitt_cost-effectiveness_1996}. However, the goal of annual screening for all diabetic patients represents a huge burden on ophthalmologists and it is far from being achieved \citep{benoit_eye_2019}. In order to improve DR screening programs, numerous Artificial Intelligence (AI) systems were thus developed to automate DR diagnosis using CFP \citep{ting_deep_2019}. However, due to the ``black-box'' nature of state-of-the-art AI, these systems still need to gain the trust of clinicians and patients.

To gain this trust, one solution investigated by \citet{araujo_drgraduate_2020} and \citet{ayhan_expert-validated_2020} is to design AI systems able to reliably estimate the uncertainty level of their predictions. This feature is expected to help clinicians know when AI predictions should be carefully reviewed and when they can be trusted. Another solution, investigated by \citet{abramoff_improved_2016}, is to develop two-stage AI systems that 1) learn to detect or segment lesions considered relevant by ophthalmologists (microaneurysms, exudates, etc.) and 2) base the AI predictions on these detections. Because they mimic ophthalmologists' reasoning, clinicians are more likely to adopt them. A similar approach was investigated by \citet{fauw_clinically_2018} for the classification of optical coherence tomography images. However, these approaches cannot generalize easily to new imaging modalities or new decision problems, such as DR progression prediction \citep{arcadu_deep_2019}, since relevant patterns are not fully known to ophthalmologists. Alternatively, another solution investigated by \citet{QuellecDeepimagemining2017a} and \citet{sayres_using_2019} is to help clinicians interpret AI predictions by highlighting image regions supposedly involved in AI predictions. If clinicians agree with highlighted areas, they will more likely trust the AI and eventually adopt it. Note that a similar approach was recently investigated for medical image segmentation \citep{wickstrom_uncertainty_2020}. However, these visualization methods provide limited information: they tell us which pixels seem to play a role in the decision process, but they do not tell us precisely how. Although interpretability is an interesting feature, it may not be enough to gain the trust of clinicians. And it is certainly not enough to gain the trust of patients, which would rather have an explanation.

\citet{gilpin_explaining_2018} differentiate interpretability and explainability as follows: \textit{interpretability} is the science of comprehending what a model did or might have done, while \textit{explainability} is the ability to summarize the reasons for an AI behavior. Explainability implies interpretability, but the reverse is not always true. To gain the trust of patients and clinicians, explainability is desirable. EXplanatory Artificial Intelligence (XAI) is a growing field of research \citep{gilpin_explaining_2018} motivated by potential AI users, worried about safety \citep{russell_research_2015}. It is also pushed by European regulations and others: the goal is to grant users the right for an explanation about algorithmic decisions that were made about them \citep{goodman_european_2017}. \added{In particular, explainability is an important consideration when clearing autonomous diagnostic AI products \citep{amann_explainability_2020, abramoff_pivotal_2018}}. An Explanatory AI system, called \explain, is presented and evaluated in this paper. Unlike visualization methods above \citep{QuellecDeepimagemining2017a, sayres_using_2019}, \explain~does not attempt to retrospectively analyze a complex classification process. Instead, it modifies the classification process in such a way that it can be understood directly.

``Black-box'' image classification AI algorithms are usually defined as Convolutional Neural Networks (CNNs) or, more generally, as ensembles of multiple CNNs \citep{ting_artificial_2019, quellec_instant_2019}. Each of these CNNs is supervised at the image level: given an image, one or several experts indicate which labels should be assigned to this image. To enable explainability, we propose to include a pixel-level classification step into the neural network. Pixel-level classification, also known as image segmentation, is generally performed by an Encoder-Decoder Network (EDN) \citep{ronneberger_u-net:_2015, lin_feature_2017, conze_abdominal_2020}. EDNs, however, are supervised at the pixel level: given an image, one or several experts assign a label to each pixel in the image. \explain~bridges the gap between the two paradigms: pixel-level classification and image-level classification are trained simultaneously, using image-level supervision only.

\added{DR severity assessment is an ideal task for evaluating \explain. In the International Clinical Diabetic Retinopathy (ICDR) severity scale \citep{wilkinson_proposed_2003}, for instance, the severity level directly derives from the abnormalities observed in CFP images: the relationship between pixel-level and image-level classifications is known. We took advantage of this property for evaluation purposes. First, we trained \explain~to automate the ICDR scale at the image level. Next, we checked whether or not 1) pixel-level classification and 2) rules linking pixel-level and image-level classifications are consistent with the ICDR scale.}

The paper is organized as follows. Related machine learning frameworks are presented in section \ref{sec:RelatedMachineLearningFrameworks}. The proposed \explain~solution is described in section \ref{sec:ExplanatoryArtificialIntelligence}. This framework is applied to DR diagnosis in section \ref{sec:Experiments}. We end up with a discussion and conclusions in section \ref{sec:DiscussionConclusions}.

\section{Related Machine Learning Frameworks}
\label{sec:RelatedMachineLearningFrameworks}

In terms of purpose, \explain~is related to existing algorithms for visualizing/interpreting what image classification CNNs have learnt. Given a trained classification CNN and an input image, these algorithms compute the influence of each pixel on CNN predictions. In the occlusion method, patches are occluded in the input image and the occluded image is run through the CNN: a drop in classification performance indicates that the occluded patch is relevant \citep{zeiler_visualizing_2014}. In sensitivity analysis, the backpropagation algorithm is used to compute the gradient of CNN predictions with respect to the value of each input pixel \citep{simonyan_deep_2014}. Various improvements on sensitivity analysis, including layer-wise relevance propagation \citep{bach_pixel-wise_2015}, also rely on back-propagated quantities to build a high-resolution heatmap showing the relevance of each pixel \citep{samek_evaluating_2017, QuellecDeepimagemining2017a}. Next, Class Activation Mapping (CAM) was proposed for image classification CNNs containing a Global Average Pooling (GAP) layer after the last convolutional layer: CAM computes CNN predictions for each of the GAP's input locations rather than for the GAP's output, thus providing coarse-resolution class-specific activation maps \citep{zhou_learning_2016}. Grad-CAM generalizes this idea to any classification CNN architecture \citep{selvaraju_grad-cam_2017}. These visualization methods attempt to retrospectively analyze a complex classification process. Another approach, investigated in this paper, is to replace the classification process with one that can be understood directly.

Because deep neural networks are hard to interpret, several authors have proposed to train deep neural decision trees \citep{yang_deep_2018} or deep neural decision forests \citep{kontschieder_deep_2016, hehn_end--end_2020} instead. These architectures are indeed based on rules that can be interpreted more easily by humans. \citet{frosst_distilling_2017} thus proposed a training procedure to derive a deep neural decision tree from a deep neural network, so that it can be interpreted. However, understanding a deep decision tree is still a challenging task for machine learning agnostics \citep{hehn_end--end_2020}: such an algorithm is interpretable, but not explainable. To enable explanations, the solution investigated in \explain~is rather to base the classification process on a segmentation and a categorization of the pathological signs, obtained solely through weak supervision.

In that sense, \explain~is also related to Weakly-Supervised Semantic Segmentation (WSSS). In WSSS, the goal is to predict pixel classes using image labels only for supervision. In particular, no positional information, like manual segmentations or bounding boxes, is required for supervision. Note that segmentation is the end goal of WSSS, while it is an intermediate step in \explain. WSSS solutions can be classified into four categories \citep{chan_comprehensive_2019}. (1) Expectation-Maximization solutions use image annotations to initialize prior assumptions about the class distribution in images. Next, an EDN is trained to meet those constraints. Then, the prior assumption model is updated based on the EDN features, and the training cycle is repeated again until convergence \citep{papandreou_weakly-_2015, pathak_constrained_2015, kervadec_constrained-cnn_2019}. (2) Multiple-Instance Learning (MIL) solutions train an image classification CNN with image-level supervision and then infer the image locations responsible for each class prediction: inference relies on the MIL assumption that an image belongs to one class if and only if at least one of its pixels does \citep{shimoda_distinct_2016, durand_wildcat_2017}. (3) Self-supervised learning solutions train an image classification CNN with image-level supervision to obtain a coarse-resolution CAM. Next, a segmentation EDN is trained using the CAM as ground truth to obtain a higher-resolution segmentation \citep{kolesnikov_improving_2016, zhang_reliability_2020}. Interestingly, the Reliable Region Mining (RRM) solution by \citet{zhang_reliability_2020} merges the two steps (CAM computation and segmentation) into one. (4) In a final category, object proposals are extracted first and the most probable class is assigned to each of them, using coarse-resolution CAMs obtained as above \citep{kwak_weakly_2017, zhou_weakly_2018, laradji_where_2019}. The self-supervised learning approach seems to be the most popular nowadays \citep{ahn_weakly_2019, lee_ficklenet_2019, jiang_integral_2019, zhang_reliability_2020, wang_self-supervised_2020}. It should be noted that \explain~is more general than WSSS solutions since the number of pixel labels can be different from the number of image labels; in particular, it can be larger. As a result, \explain~is more semantically rich and it can lead to better image classification, as it allows for more general ``pixel to image'' label inference rules.

Note that the proposed framework is related to the GP-Unet solution by \citet{dubost_weakly_2020}: both solutions include an EDN inside an image classification CNN, in order to generate high-resolution attention maps. \explain~differs in that it generates multiple, complementary maps: one map is generated per type of discriminant patterns in images. It also differs in that image classification derives very simply from the pixel classifications in order to allow explainability, rather than interpretability. Finally, \explain~introduces a new criterion, namely the generalized occlusion method, in order to optimize foreground/background separation in the pixel classification maps.

\section{Explanatory Artificial Intelligence}
\label{sec:ExplanatoryArtificialIntelligence}

\subsection{Overview and Notations}
\label{sec:OverviewNotations}

\begin{figure*}[t]
  \begin{center}
    \includegraphics[width=\textwidth]{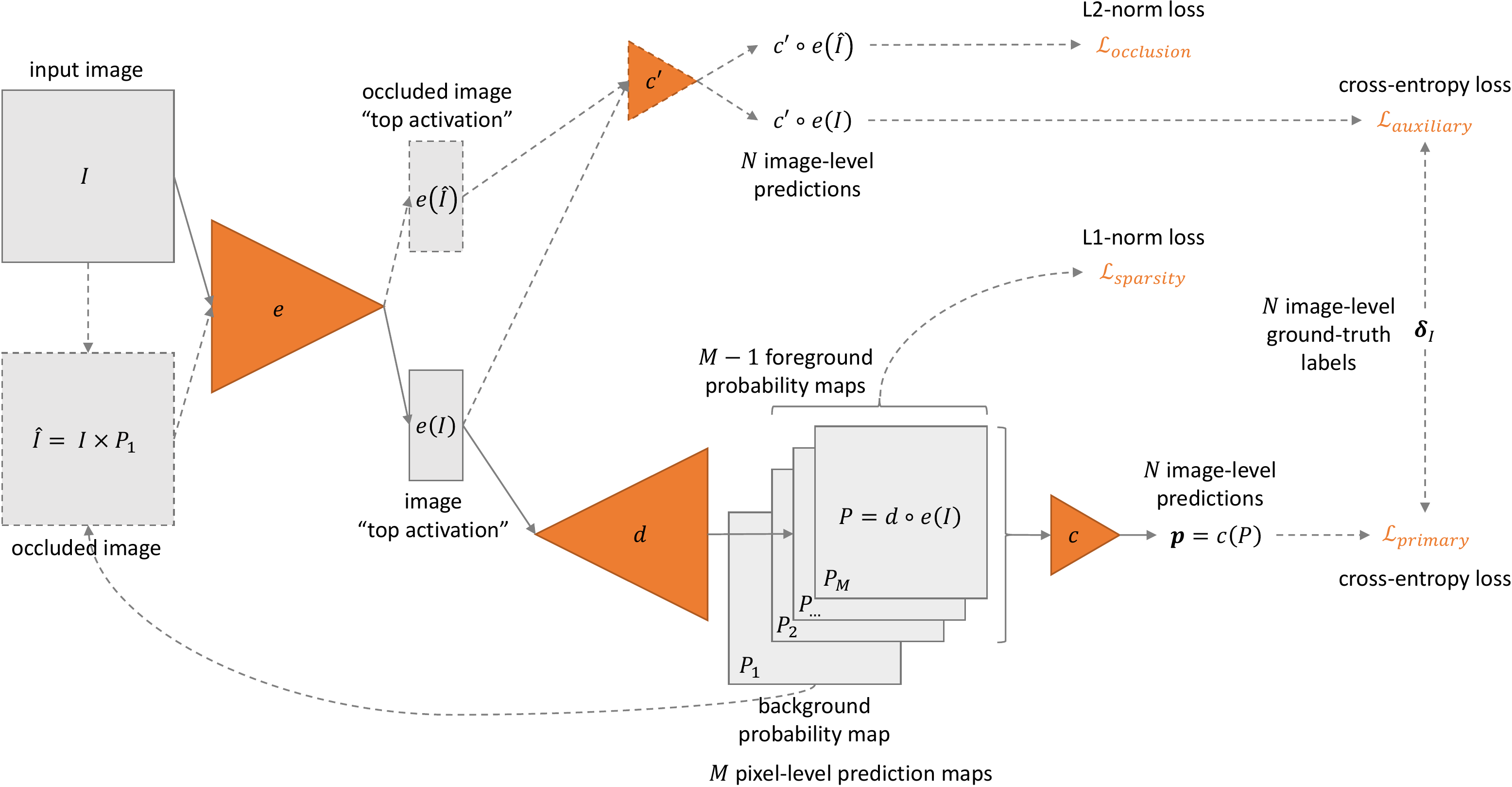}
  \end{center}
  \caption{Outline of the \explain~framework. Dashed lines indicate operations performed during training only.}
  \label{fig:outline}
\end{figure*}

This paper addresses multilabel image classification: given an input image $I$ and $N$ image-level labels, the goal is to predict whether or not experts would assign the $n$-th label to image $I$, $\forall n \in \{1, ..., N \}$. Let $p_n \in \left[0; 1\right]$ denote the probabilistic prediction of \explain~and let $\delta_{I,n} \in \{0,1\}$ denote the ground truth: did experts actually assign the $n$-th label to image $I$?  Unlike multiclass classification, multilabel classification does not assume image-level labels to be mutually exclusive: $\sum_{n=1}^N{\delta_{I,n}} \in \{ 0, 1, ..., N \}$. Images where $\sum_{n=1}^N{\delta_{I,n}}=0$ are referred to as ``background images'', i.e. images where experts did not annotate anything.

As an intermediate step, \explain~also assigns a label to each (color or grayscale) pixel $I_{x,y}$ for explainability purposes. Let $M$ denote the number of pixel-level labels:
\begin{itemize}[noitemsep]
  \item $M$ can be smaller than, equal to, or larger than the number $N$ of image-level labels,
  \item the first of these pixel-level labels represents ``background pixels''.
\end{itemize}
A pixel-level prediction tensor $P$ is thus calculated, where $P_{m,x,y} \in \left[0; 1\right]$ indicates the probability that pixel $I_{x,y}$ should be assigned the $m$-th pixel-level label, with $m \in \{ 1, ..., M \}$. Unlike image-level labels, pixel-level labels are assumed to be mutually exclusive: $\sum_{m=1}^M{P_{m,x,y}}=1$ (multiclass classification). Since our framework is trained with image supervision only, we assume that no ground truth is available for pixel-level labels.

As illustrated in Fig.~\ref{fig:outline}, the pixel-level and image-level classification problems are solved jointly as follows:
\begin{itemize}[noitemsep]
  \item an Encoder-Decoder Network (EDN) $s$ predicts $P$ from $I$,
  \item a classification head $c$ predicts $\boldsymbol{p}=\left\lbrace p_n, \forall n \in \{ 1, ..., N \} \right\rbrace$ from $P$,
  \item additional branches are included during training, to compute auxiliary losses improving explainability.
\end{itemize}

\subsection{Pixel-Level Label Prediction}
\label{sec:PixelLevelLabelPrediction}

In order to predict pixel-level labels $P_{m,x,y}$ for each pixel $I_{x,y}$, an EDN $s$ is used: $P = s(I)$. This EDN is composed of an encoder network $e$ and a decoder network $d$ such that $s = d\circ e$.

The encoder part $e$ of $s$ was defined as an EfficientNet \citep{tan_efficientnet_2019}. EfficientNet is a family of CNN architectures of increasing complexity. The smallest model, namely EfficientNet-B0, was obtained through neural architecture search. Larger models, EfficientNet-B1 to -B7, were then obtained by up-scaling EfficientNet-B0: depth, width and resolution were increased proportionally. Although any classification CNN may be used as backbone for $s$, EfficientNets were chosen for their clearly superior tradeoff between accuracy and complexity, compared to traditional deep networks.

The overall architecture of $s$ was defined as a Feature Pyramid Network (FPN) by \citet{lin_feature_2017}. Like the well-know U-Net \citep{ronneberger_u-net:_2015}, an FPN is a top-down architecture with skip-connections between $e$ and $d$. Unlike U-Net, high-level semantic feature maps are produced at multiple scales: these multi-scale feature maps are then 1) up-scaled to match the size of $I$, 2) concatenated and 3) processed by a final convolutional layer to obtain the output tensor $P$, following a deep supervision strategy \citep{xie_holistically-nested_2015}. Although any EDN architecture may be used, FPN was chosen for its faster convergence.

Because pixel-level labels are mutually exclusive, a softmax operator $\mu$ was used as activation function for the last convolutional layer:
\begin{equation}
  \label{eq:softmax}
  \mu(\boldsymbol{z}) = \left\lbrace \frac{e^{z_m}}{\sum_{i=1}^M{e^{z_i}}}, \forall m \in \{ 1, ..., M\} \right\rbrace \;,
\end{equation}
where $\boldsymbol{z}$ are the outputs of the $M$ neurons at a given pixel location.

\subsection{Image-Level Label Prediction}
\label{sec:ImageLevelLabelPrediction}

In order to predict the vector $\boldsymbol{p}$ of image-level labels from the tensor $P$ of pixel-level labels, a classification head $c$ is used: $\boldsymbol{p} = c(P) = c\circ s(I)$. $c$ has to be very simple to enable explainability. In practice, it consists of two layers only.

\subsubsection{Summary Layer}
The first layer, $\Pi$, summarizes each label prediction map $P_m$ by a single scalar value. Two easily understandable summaries may be considered:
\begin{enumerate}[noitemsep]
  \item the average value, which is proportional to the surface covered by each pixel-level label,
  \item the maximal value, which represents the strongest clue of presence for each pixel-level label.
\end{enumerate}
In theory, both solutions enable explainability. However, in practice, the first option has one major drawback: it favors oversegmentation in foreground images, which limits explainability. Indeed, to improve separation between background and foreground images, the EDN is encouraged to increase the average predictions for foreground pixel-level labels in foreground images (and decrease them in background images): one way is to increase the number of foreground pixels in foreground images, i.e. to oversegment. \added{In the second option, the maximal value is independent from the number of foreground pixels. Therefore, this option does not favor over- or under-segmentation:} \changed{this option was selected.} $\Pi$ is thus defined as the global max pooling layer:
\begin{equation}
  \label{eq:summaryLayer}
  \Pi(P) = \left\lbrace \max_{x,y}{P_{m,x,y}}, \forall m \in \{ 2, ..., M \} \right\rbrace \;.
\end{equation}
Note that the background probability map $P_1$ is not used for image-level prediction.

\subsubsection{Classification Layer}
The second layer, $\Delta$, is a special class of dense layers, where neural weights are all positive. The positivity constraint improves explainability: the image-level prediction is defined as a sum of pixel-level clues, each clue being associated with a (positive) confidence level. Layer $\Delta$ is defined as follows:
\begin{equation}
  \label{eq:deltaLayer}
  \Delta(\boldsymbol{z}) = \left\lbrace \sigma \left( \sum_{m=2}^M{z_m w_{m,n}^2} + b_n \right), \forall n \in \{ 1, ..., N \} \right\rbrace \;,
\end{equation}
where $w_{m,n}^2$ are positive neural weights, $b_n$ are biases and $\sigma$ is an activation function. Because image-level labels are not mutually exclusive, $\sigma$ was defined as a sigmoid function:
\begin{equation}
  \label{eq:sigmoid}
  \sigma(z) = \frac{1}{1+\exp(-z)} \;.
\end{equation}

\subsection{Learning to Detect Background Pixels}
\label{sec:LearningDetectBackgroundPixels}

In \explain, pixel-level labels are not necessarily related to image-level labels. However, for localization purposes, ``background pixels'' are related to ``background images''. We remind that: 1) background pixels are those associated with the first pixel-level label and 2) background images are those associated with no image-level label ($\sum_{n=1}^N{\delta_{I,n}}=0$). To improve explainability, we propose that background images only contain background pixels.

To ensure this property, a generalization of the occlusion method \citep{zeiler_visualizing_2014} is proposed. The original occlusion method generates multiple occluded versions of the input image $I$ by zeroing all pixel intensities inside a sliding square window. These occluded versions are then run through a previously trained CNN in order to detect background (or conversely foreground) pixels. Instead, we propose to generate a single occluded version $\hat{I}$ of the input image. Let $P_1 = \left\lbrace P_{1,x,y}, \forall (x, y) \right\rbrace$ denote the background probability map. The occluded image $\hat{I}$ is defined as follows:
\begin{equation}
  \hat{I} = I \times P_1 = \left\lbrace I_{x,y} P_{1,x,y}, \forall (x, y) \right\rbrace \;,
\end{equation}
where $\times$ is the element-wise product. If $I$ is a color image, each color component of pixel $I_{x,y}$ is multiplied by the same value $P_{1,x,y}$. Unlike the original occlusion method, this generalized occlusion method is performed during training, in order to optimize the background probability map $P_1$. The following two properties are optimized jointly:
\begin{description}[noitemsep]
  \item[Occlusion sensitivity:] the occluded image $\hat{I} = I\times P_1$ should always be perceived as a background image, regardless of the ground-truth image-level labels. This indicates that all relevant pixels have been successfully occluded: occlusion is sensitive.
  \item[Occlusion specificity:] the background $P_1$ should be as extended as possible or, conversely, the foreground image $1 - P_1$ should be as sparse as possible. This indicates that occlusion is specific to relevant pixels.
\end{description}

\subsection{Auxiliary Classification Branch}
\label{sec:AuxiliaryClassificationBranch}

In order to optimize the first property above, namely occlusion sensitivity, the occluded image $\hat{I} = I\times P_1$ should run through a classifier, and the background image $P_1$ should be optimized in such a way that $\hat{I}$ is predicted as a background image. The $c\circ s$ classifier may be used for that purpose. However, optimizing $c\circ s(\hat{I})$ would not only alter background pixel detection, it will potentially alter the entire classifier. Therefore, an independent classification branch should be used instead of $c\circ s$.

For the related problem of CAM computation in WSSS, \citet{zhang_reliability_2020} used a completely independent classifier. However, we assume that foreground/background separation is mainly performed by the decoder part $d$ of the EDN $s = d\circ e$ (see section \ref{sec:PixelLevelLabelPrediction}). Therefore, we propose to reuse the encoder part $e$ of $s$ for this auxiliary classification branch. It has the advantage of significantly reducing training complexity, while alllowing a more generic feature extraction. In that purpose, an auxiliary classification head $c'$ is connected to the $L$-channel tensor $T = e(I)$, i.e. to the ``top activation'' layer of $s$. Following common practice, this classification head consists of a global average pooling layer, followed by a regular dense layer. Like $c$, this classification head has $N$ non-mutually exclusive outputs:
\begin{equation}
  c'(T) = \left\lbrace \sigma\left( \sum_{l=1}^L{  \frac{ \sum_{x,y}{T_{l,x,y}} }{ \sum_{x,y}{1} } w'_{l,n}} + b'_n \right), \forall n \in \{ 1, ..., N \} \right\rbrace \;,
\end{equation}
where $\sigma$ is the sigmoid function of Eq. (\ref{eq:sigmoid}), and where $w'_{l,n}$ and $b'_n$ are neural weights and biases, respectively.

To summarize, $c' \circ e$ is the classification branch used to classify occluded images, for the purpose of optimizing background images.

\subsection{Loss Functions}
\label{sec:LossFunctions}

The entire neural architecture has been described. This section enumerates the loss functions to optimize both pixel-level and image-level classification performance.

\subsubsection{Cross-Entropy Losses}

The main goal of the proposed framework is to correctly classify image-level labels. Given network predictions $\boldsymbol{p}=c\circ s(I)$ and ground truth labels $\boldsymbol{\delta}_I = \left\lbrace \delta_{I,n}, \forall n \in \{ 1, ..., N \} \right\rbrace$, the primary loss function $\mathcal{L}_{primary}$ is thus defined as a cross-entropy loss function:
\begin{equation}
  \label{eq:crossEntropy}
  \begin{array}{rl}
    \mathcal{L}_{primary} = -\frac{\displaystyle 1}{ N \displaystyle \sum_I{1} }
      \displaystyle\sum_I\sum_{n=1}^N & \left[\delta_{I,n}\log(c\circ s(I)_n) + \right. \\
                                      & \left. (1-\delta_{I,n})\log(1 - c\circ s(I)_n) \right] \;,
  \end{array}
\end{equation}
where $c\circ s(I)_n = p_n, n \in \{ 1, ..., N \}$.

Because an auxiliary classification branch $c'$ is defined in section \ref{sec:AuxiliaryClassificationBranch}, $N$ auxiliary image-level predictions $\boldsymbol{p}'=c'\circ e(I)$ should also be optimized. An auxiliary loss function $\mathcal{L}_{auxiliary}$ is thus defined similarly to $\mathcal{L}_{primary}$, by replacing $c\circ s$ with $c'\circ e$ in equation \ref{eq:crossEntropy}.

\subsubsection{Occlusion Loss}

An additional loss function $\mathcal{L}_{occlusion}$ is defined to optimize the first property of the background probability map $P_1$, namely that the occluded image $\hat{I} = I\times P_1$ should always be perceived as a background image (see section \ref{sec:LearningDetectBackgroundPixels}). Given the definition of a background image, namely $\sum_{n=1}^N{\delta_{I,n}}=0$, $\mathcal{L}_{occlusion}$ is defined as the squared L2-norm of $c'\circ e \left( \hat{I} \right)$ predictions:
\begin{equation}
  \label{eq:occlusion}
  \mathcal{L}_{occlusion} = \frac{ \displaystyle \sum_I \sum_{n=1}^N{ \left[ c'\circ e \left( \hat{I} \right)_n \right]^2 } }{ N \displaystyle \sum_I{1} } \;.
\end{equation}

\subsubsection{Sparsity Loss}

A final loss function $\mathcal{L}_{sparsity}$ is defined to optimize the second property of the background probability map $P_1$, namely that the foreground map $1 - P_1$ should be sparse (see section \ref{sec:LearningDetectBackgroundPixels}). $\mathcal{L}_{sparsity}$ is defined as the L1-norm of $1 - P_1$ maps. Because a softmax activation function is used at the end of $s$, $\mathcal{L}_{sparsity}$ can be expressed as follows (see Eq. (\ref{eq:softmax})):
\begin{equation}
  \mathcal{L}_{sparsity} = \frac{ \displaystyle \sum_I \sum_x \sum_y{ \left| \sum_{m=2}^M{ s(I)_{m,x,y} } \right| } }{ \displaystyle \sum_I \sum_x \sum_y{1} } \;.
\end{equation}
Note that the modulus operator can be dropped since $0 \leq \sum_{m=2}^M{ s(I)_{m,x,y} } \leq 1, \forall (x, y)$, due to the use of a softmax activation function in section \ref{sec:PixelLevelLabelPrediction}.

\subsubsection{Total Loss}

Those four loss functions are combined linearly to obtain the total loss function $\mathcal{L}$ that should be minimized during training:
\begin{equation}
  \label{eq:totalLoss}
  \mathcal{L} = \mathcal{L}_{primary} + \alpha \mathcal{L}_{auxiliary} + \beta \mathcal{L}_{occlusion} + \gamma \mathcal{L}_{sparsity} \; ,
\end{equation}
where $\alpha \geq 0$, $\beta \geq 0$ and $\gamma \geq 0$.

\subsubsection{Loss Function Competition}
\label{sec:LossFunctionCompetition}

Ideally, convergence to a suitable classifier should not depend critically on the choice of $\alpha$, $\beta$ and $\gamma$ weights. In other words, the four basic loss functions should not compete with one other.

First, $\mathcal{L}_{auxiliary}$ does not compete with $\mathcal{L}_{primary}$: whether we want to segment and classify pathological signs (through $\mathcal{L}_{primary}$) or directly classify images without segmentation (through $\mathcal{L}_{auxiliary}$), the objective of the shared encoder $e$ is to encode the presence of pathological signs in images.

$\mathcal{L}_{occlusion}$ and $\mathcal{L}_{sparsity}$, on the other hand, are competing. By design, these loss functions optimize occlusion sensitivity and specificity (see section \ref{sec:LearningDetectBackgroundPixels}), two metrics that always need to be traded off. $\mathcal{L}_{sparsity}$ and $\mathcal{L}_{primary}$ are also competing: if all pixels are assigned to the background (high foreground sparsity), then no pathological sign can be detected and the classifier will always predict background images, leading to poor classification performance. Fortunately, assuming that decoder $d$ ensures foreground/background separation, $\mathcal{L}_{auxiliary}$ and $\mathcal{L}_{sparsity}$ are independent. Therefore, encoder $e$, the largest part of the network, can always be trained through $\mathcal{L}_{auxiliary}$, even if the foreground is temporarily too sparse. Besides training complexity, this property also motivates the use of a partly independent classification branch for occlusion sensitivity maximization, over a completely independent classification branch (see section \ref{sec:AuxiliaryClassificationBranch}).

Finally, $\mathcal{L}_{occlusion}$ has no reason to compete with $\mathcal{L}_{primary}$ and $\mathcal{L}_{auxiliary}$. In conclusion, $\gamma$, the weight controlling $\mathcal{L}_{sparsity}$, is the most critical weight. It can be used to trade-off image-level classification performance and pixel-level classification quality (i.e. explainability).

\subsection{Explanation Generation}
\label{sec:ExplanationGeneration}

Once training has converged, the proposed system can be used to infer automatic diagnoses for an unseen image $I$. In addition to $N$ image-level predictions, $M-1$ pixel-level probability maps are obtained (see Fig. \ref{fig:outline}). The following procedure is proposed to explain automatic diagnoses for $I$:
\begin{enumerate}[noitemsep]
  \item The $M-1$ pixels maximizing pixel-level prediction in each of the $M-1$ foreground maps can be highlighted in $I$ (see Eq. (\ref{eq:summaryLayer})).
  \item Let $i_2$, $i_3$, ..., $i_M$ denote the (positive) intensity of those pixels. To explain the $n$-th image-level decision, those intensities should be multiplied, respectively, by the (positive) weights $w^2_{2,n}$, $w^2_{3,n}$, ..., $w^2_{M,n}$ involved in $c$ (see Eq. (\ref{eq:deltaLayer})). Product $i_m w^2_{m,n}$ indicates the strength of clue number $m$ in the $n$-th decision.
  \item Additionally, if experts can associate a keyword (typically lesion names) with each pixel-level label, then the previous step can be converted into a set of sentences.
\end{enumerate}

\section{Application to Diabetic Retinopathy Diagnosis}
\label{sec:Experiments}

The proposed framework has been applied to the diagnosis of DR through the analysis of CFP images. The goal was to grade DR severity in one eye according to the \changed{ICDR} scale \citep{wilkinson_proposed_2003}:
\changed{\begin{description}[noitemsep]
  \item[No apparent DR --] no abnormalities.
  \item[Mild nonproliferative DR (NPDR) --] microaneurysms only.
  \item[Moderate NPDR --] more than just microaneurysms but less than severe NPDR.
  \item[Severe NPDR --] any of the following: more than 20 intraretinal hemorrhages in each of 4 quadrants; definite venous beading in 2 quadrants; prominent intraretinal microvascular abnormalities in 1 quadrant and no signs of proliferative DR (PDR).
  \item[PDR --] one or more of the following: neovascularization, vitreous/preretinal hemorrhage.
\end{description}
By design, this multiclass classification problem can be transformed} into a four-label multilabel classification problem ($N=4$): `at least mild NPDR', `at least moderate NPDR', `at least severe NPDR' and `PDR'. Each of these labels \changed{is associated with the presence of specific abnormalities (``microaneurysms'' for `at least mild NPDR', ``more than just microaneurysms'' for `at least moderate NPDR', etc.)}. Note that 1) background images correspond to `no apparent DR' and that 2) detecting the `no apparent DR' is the exact opposite of detecting `at least mild NPDR'.

\subsection{Image Datasets}

A model was developed using training and validation datasets with image-level labels. Next, this model was evaluated on \changed{multiple datasets with image-level labels and on one dataset with pixel-level labels}. \added{In all datasets, images were graded according to the ICDR scale.} In practice, image-level labels were not assigned to single images but, more generally, to small sets of images (CFPs) from the same eye. These datasets have various origins: USA, France, India \added{and China}.

\subsubsection{OPHDIAT Image-Level Evaluation Dataset}

\changed{The first image-level evaluation dataset} originates from the OPHDIAT DR screening network, which consists of 40 screening centers located in 22 diabetic wards of hospitals, 15 primary health-care centers and 3 prisons in the Ile-de-France area \citep{massin_ophdiat:_2008}. \added{Let $OPHDIAT_{eval}$ denote this dataset.} This dataset consists of 21,576 CFPs from 9,734 eyes of 4,996 patients ($\sim$ 2 CFPs per eye). As part of the quality-assurance program of OPHDIAT, DR severity in these randomly-selected patients was graded by two ophthalmologists. In case of disagreement, images were read a third time by a senior ophthalmologist.

\subsubsection{Training and Validation Datasets}

The training and validation datasets originate from two DR screening programs: \changed{OPHDIAT and EyePACS \citep{cuadros_eyepacs:_2009}}.

Images from EyePACS were acquired in multiple primary care sites throughout California and elsewhere. A total of 88,702 CFPs from 88,702 eyes of 44,351 patients (one CFP per eye), released for Kaggle's Diabetic Retinopathy Detection challenge\footnote{\url{https://www.kaggle.com/c/diabetic-retinopathy-detection/data}}, was used. Training and validation images from OPHDIAT consist of all images from patients that were not included in the image-level evaluation dataset: a total of 610,748 CFPs from 275,236 eyes of 142,145 patients ($\sim$ 2 CFPs per eye) were included. In both subsets, DR severity in each eye was graded by a single human reader.

Images were distributed as follows: 90\% were used for training and 10\% were used for validation. The same proportion of eyes from EyePACS and OPHDIAT (24.4\% / 75.6\%) was used in the training and validation datasets.

\subsubsection{IDRiD \changed{Image- and Pixel-Level Evaluation Datasets}}

\changed{Performance was also evaluated on the Indian Diabetic Retinopathy Image Dataset (IDRiD)\footnote{\url{https://idrid.grand-challenge.org}}, which originates from an eye clinic located in Nanded, India. This dataset was collected for the purpose of an image analysis challenge \citep{porwal_idrid:_2019}.} \added{For image-level evaluation, we used the 103 CFPs released for the onsite evaluation of algorithms competing in sub-challenge 2 (disease grading). Let $IDRiD_{image}$ denote this dataset.} \changed{For pixel-level evaluation, we used the 143 CFPs released for training the algorithms competing in sub-challenge 1 (lesion segmentation). Let $IDRiD_{pixel}$ denote this dataset.} Experts manually segmented four types of abnormalities related to DR in those images: microaneurysms, hemorrhages, soft exudates and hard exudates \citep{porwal_idrid:_2019}. For each CFP, one binary segmentation map is thus available per lesion type. \added{No training or fine-tuning was performed on IDRiD data, to ensure independent evaluation.}

\subsubsection{\added{DeepDR Image-Level Evaluation Dataset}}

\added{Finally, image-level performance was evaluated on images from the DeepDR challenge\footnote{\added{\url{https://isbi.deepdr.org}}}. We used the 400 CFPs from 200 eyes of 100 patients (2 CFPs per eye) released for validating the algorithms competing in sub-challenge 1 (disease grading). Part of these images originate from screening programs in China: the Shanghai Diabetic Complication Screening Project (SDCSP), the Nicheng Diabetes Screening Project (NDSP), and Nationwide Screening for Complications of Diabetes (NSCD). Some images originate from an eye clinic located in the Department of Ophthalmology, Shanghai Jiao Tong University, China. Let $DeepDR_{valid}$ denote this dataset. The onsite evaluation dataset, denoted by $DeepDR_{onsite}$, is not publicly-available yet. No training or fine-tuning was performed on DeepDR data, to ensure independent evaluation.}

\subsection{Image Pre-processing}

The image pre-processing procedure proposed by B. Graham for the ``min-pooling'' solution, which ranked first in the Kaggle Diabetic Retinopathy competition, was followed.\footnote{\url{https://www.kaggle.com/c/diabetic-retinopathy-detection/discussion/15801}} First, to focus the analysis on the camera's field of view, images were adaptively cropped and resized to 448$\times$448 pixels. Second, to attenuate intensity variations throughout and across images, image intensity was normalized. Normalization was performed in each color channel independently: the background was estimated using a large Gaussian kernel and subtracted from the image \citep{QuellecDeepimagemining2017a}.

Data augmentation was performed during training. Before feeding a pre-processed image to a CNN, the image was randomly rotated (range: [0$^{\circ}$, 360$^{\circ}$]), translated (range: [-15\%, 15\%]), scaled (range: [80\%, 120\%]) and horizontally flipped. Additionally, Gaussian noise (standard deviation: 5\% of the intensity range) was added and the contrast was randomly scaled (range: [75\%, 125\%]).

\subsection{Managing Eye-Level Labels}

In OPHDIAT, which is used for training, validation and image-level evaluation, labels (DR severity) are assigned to eyes. To facilitate training, a single image was selected per eye at the beginning of each epoch: the one maximizing the sum of image-level predictions ($\sum_{n=1}^N{p_n}$), i.e. the most pathological image of each eye. For image-level evaluation, the maximal prediction among images of the same eye was considered for each label $n \in \{ 1, ..., N \}$.

\subsection{Baseline Methods}

\subsubsection{Reliable Region Mining (RRM)}

The self-supervised WSSS solution by \citep{zhang_reliability_2020}, named Reliable Region Mining (RRM), was used as baseline. Like the proposed solution, RRM jointly trains a classification branch and a segmentation branch. The purpose of the classification branch in RRM is to compute the CAMs, which are used to supervise the segmentation branch, after post-processing by a dense Conditional Random Field (CRF). However, unlike the proposed solution, segmentations are not used for classification, so only segmentation results can be compared. RRM also use FPN as EDN architecture. For a fair comparison, the same family of classification CNNs (EfficientNet) was used as backbones.

\subsubsection{\explain-CAM}

In order to evaluate the proposed occlusion strategy (see section \ref{sec:LearningDetectBackgroundPixels}), a variation on \explain~and RRM, called \explain-CAM, was also evaluated. In \explain-CAM, the background probability map $P_1$ is not trained to maximize occlusion sensitivity and specificity. Instead, what is maximized is the Dice similarity coefficient between the foreground probability map ($1-P_1$) and the element-wise maximum of class-specific CAMs. The CAMs are computed as in RRM, using the classification branch.

\subsubsection{``Black-Box'' AI}

Finally, in order to evaluate the impact of explainability on image-level performance, \explain~was compared to a previous ``black-box'' AI solution from our group, relying on an ensemble of multiple CNNs \citep{quellec_instant_2019}.

\subsection{\added{Implementation Details}}

\added{\explain~was implemented using the \textit{Keras} API of the \textit{TensorFlow} library\footnote{\added{\url{https://www.tensorflow.org}}}. Encoder-Decoder Networks (EDNs) were based on the \textit{Segmentation Models} library \footnote{\added{\url{https://github.com/qubvel/segmentation_models}}}. The \textit{OpenCV}\footnote{\added{\url{https://github.com/opencv/opencv-python}}} and \textit{imgaug}\footnote{\added{\url{https://imgaug.readthedocs.io}}} libraries were used for image preprocessing and data augmentation, respectively.}

\subsection{Image-Level Performance}

\subsubsection{Performance Metrics}

\changed{First, we evaluated multilabel classification performance at the image level. Receiver-Operating Characteristics (ROC) curves were used}: one ROC curve was obtained per image-level label $n \in \{ 1, ..., N \}$. Each curve was obtained by varying a cutoff on the output probabilities $p_n$ and by measuring the classification sensitivity and specificity for each cutoff; ROC curves were obtained by connecting (1-specificity, sensitivity) pairs. Classification performance for one label can be summarized by the Area Under the ROC Curve (AUC).

\added{Next, we evaluated multiclass classification performance at the image level. In this scenario, one severity level $s$ must be assigned to each image: either `no apparent DR', `mild NPDR', `moderate NPDR', `severe NPDR' or `PDR'. Given four probability cutoffs $\tau_1$, $\tau_2$, $\tau_3$ and $\tau_4$, this severity level was defined as follows:
\begin{itemize}[noitemsep]
  \item if $p_4 \geq \tau_4$: $s$ = `PDR',
  \item else if $p_3 \geq \tau_3$: $s$ = `severe NPDR',
  \item else if $p_2 \geq \tau_2$: $s$ = `moderate NPDR',
  \item else if $p_1 \geq \tau_1$: $s$ = `mild NPDR',
  \item else: $s$ = `no apparent DR'.
\end{itemize}
Performance was evaluated using the accuracy and the quadratic weighted Kappa ($\kappa$), a variation on Cohen's Kappa for ordinal scales, which takes into account the amount of disagreement between observers \citep{cohen_weighted_1968}. The probability cutoffs were chosen to maximize $\kappa$ on the validation set.}

\begin{figure}[t]
  \begin{center}
    \includegraphics[width=\textwidth]{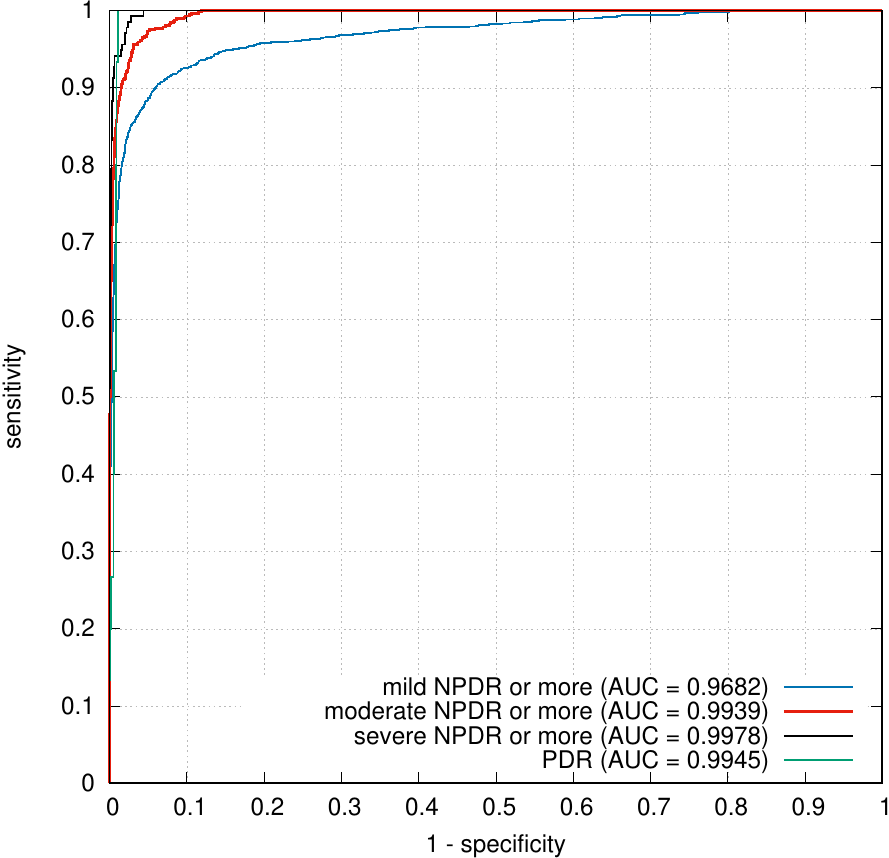}
  \end{center}
  \caption{Receiver-operating characteristics for DR severity assessment, i.e. image-level classification, in the OPHDIAT \citep{massin_ophdiat:_2008} evaluation dataset, using \changed{the EfficientNet-B5 backbone and $M=6$ pixel-level labels}.}
  \label{fig:imageLevelROC}
\end{figure}

\subsubsection{Hyperparameter Optimization}

Hyperparameters of each algorithm were optimized at the image level using the validation dataset. A ROC analysis was performed: hyperparameters maximizing the average per-label AUC on the validation subset were selected. For \explain, the following hyperparameters had to be set:
\begin{itemize}[noitemsep]
  \item the number $M$ of pixel-level labels (see section \ref{sec:OverviewNotations}),
  \item the CNN backbone (see section \ref{sec:PixelLevelLabelPrediction}),
  \item training weights $\alpha$, $\beta$ and $\gamma$ (see section \ref{sec:LossFunctions}).
\end{itemize}
Training weight $\gamma$, which trades off image-level and pixel-level classification quality (see section \ref{sec:LossFunctionCompetition}), was empirically set to $0.1$. Then, $M$, $\alpha$ and $\beta$ were chosen to maximize the average per-backbone and per-label AUC. Finally, the best CNN backbone, given the optimal $M$, $\alpha$ and $\beta$, was selected. A similar procedure was followed for each algorithm ($\gamma$ or equivalent set to 0.1, CNN backbone chosen last).

\subsubsection{Results}

\begin{table}[t]
    \caption{Image-level classification performance of various architectures, in terms of AUC, in the OPHDIAT evaluation dataset. The ``/Bx'' suffix in the architecture names indicates the EfficientNet-``Bx'' model used as backbone. \changed{The selected \explain~model, which maximizes} the mean AUC (both in the validation subset and in the OPHDIAT evaluation dataset), is in bold. EfficientNet-B1 is the best backbone for \explain-CAM in terms of mean AUC (both in the validation subset and in the OPHDIAT evaluation dataset). \added{`Combination' denotes the optimal combination of multiple \explain~models.}}
    \begin{center}
      \begin{tabular}{c|cccc}
        architecture    & \RotText{at least mild NPDR} & \RotText{at least moderate NPDR} & \RotText{at least severe NPDR} & \RotText{PDR} \\
        \hline
        \explain/B0          & 0.9500 & 0.9886 & 0.9968 & 0.9938 \\
        \explain/B1          & 0.9586 & 0.9917 & 0.9970 & 0.9905 \\
        \explain/B2          & 0.9602 & 0.9930 & 0.9971 & 0.9890 \\
        \explain/B3          & 0.9535 & 0.9906 & 0.9972 & 0.9933 \\
        \explain/B4          & 0.9634 & 0.9929 & 0.9972 & 0.9900 \\
        \textbf{\explain/B5} & \textbf{0.9682} & \textbf{0.9939} & \textbf{0.9978} & \textbf{0.9945} \\
        \explain/B6          & 0.9627 & 0.9923 & 0.9972 & 0.9945 \\
        \explain/B7          & 0.9620 & 0.9918 & 0.9941 & 0.9914 \\
        \hline
        \added{Combination}  & \added{0.9691} & \added{0.9941} & \added{0.9977} & \added{0.9958} \\
        \hline
        \explain-CAM/B1      & 0.9523 & 0.9918 & 0.9970 & 0.9874 \\
        ``black-box'' AI     & 0.9702 & 0.9859 & 0.9969 & 0.9969 \\
      \end{tabular}
    \end{center}
    \label{tab:imageLevelAUC}
\end{table}

The hyperparameters of \explain~maximizing image-level classification performance on the validation dataset are as follows:
\begin{itemize}[noitemsep]
  \item $M=6$ pixel-level labels (see section \ref{sec:OverviewNotations}),
  \item EfficientNet-B5 backbone (see section \ref{sec:PixelLevelLabelPrediction}),
  \item training weights $\alpha = \beta = 0.1$ (see section \ref{sec:LossFunctions}).
\end{itemize}

ROC curves obtained with these hyperparameters in the \changed{$OPHDIAT_{eval}$ dataset} are reported in Fig. \ref{fig:imageLevelROC}. For comparison purposes, image-level performance achieved with different EfficientNet backbones and with the \explain-CAM and ``black-box'' AI baselines is summarized in Table \ref{tab:imageLevelAUC}. \added{The combination of multiple \explain~models was also investigated: eight models with different backbones (EfficientNet-B0 to -B7) were combined through logistic regression on the validation dataset. Results on $OPHDIAT_{eval}$ are reported in Table \ref{tab:imageLevelAUC}. Combining multiple models improves image-level classification performance. However, the improvement is limited and it comes at the expense of reduced explainability. Therefore, a simple \explain~model (with the optimal hyperparameters) is used in the remainder of the study.}

\added{Next, the multiclass classification performance of this solution, on multiple evaluation datasets, is reported in Table \ref{tab:imageLevelAccKappa}: \explain~is compared with competing solutions of the IDRiD and DeepDR challenges.}

\begin{table}[t]
    \caption{\added{Image-level classification performance, in terms of accuracy and quadratic weighted Kappa ($\kappa$), in various datasets. The ``proposed'' approach is \explain~with the EfficientNet-B5 backbone and $M=6$ pixel-level labels. Results of the IDRiD challenge originate from \citet{porwal_idrid:_2019}. Results of the DeepDR challenge are from the challenge website: the 4 best solutions (out of 8) are reported (SJTU: Shanghai Jiao Tong University; CUHK: Chinese University of Hong Kong) --- the onsite evaluation dataset is not public yet.}}
    \added{
    \begin{center}
      \begin{tabular}{c|c|cc}
        dataset          & method    & accuracy               & $\kappa$ \\
        \hline
        $IDRiD_{image}$  & proposed  & 0.75 (0.7476)          & 0.8673 \\
                         & LzyUNCC   & 0.75                   &        \\
                         & VRT       & 0.59                   &        \\
                         & Mammoth   & 0.54                   &        \\
                         & HarangiM1 & 0.55                   &        \\
                         & AVSASVA   & 0.55                   &        \\
                         & HarangiM2 & 0.48                   &        \\
        \hline
        $DeepDR_{valid}$ & proposed & 0.7300                & 0.9243 \\
        \hdashline
        $DeepDR_{onsite}$\footnotemark[9]
                         & SJTU (M1) & & 0.9215 \\
                         & SJTU (M2) & & 0.9211 \\
                         & Vuno Inc. & & 0.9097 \\
                         & CUHK      & & 0.8845 \\
        \hline
        $OPHDIAT_{eval}$ & proposed  & 0.9389                & 0.8937 \\
      \end{tabular}
    \end{center}
    }
    \label{tab:imageLevelAccKappa}
\end{table}
\footnotetext[9]{\added{\url{https://isbi.deepdr.org/leaderboard.html}}}

\subsection{Pixel-Level Performance}

\subsubsection{Performance Metrics}
\label{sec:QuantitativePixelLevelEvaluation}

Classification performance at the pixel level can also be evaluated by ROC curves. However, because positive and negative pixels are highly unbalanced, a second evaluation method was used: Precision-Recall (PR) curves. PR curves are built similarly to ROC curves, but (recall, precision) pairs are used in place of (1-specificity, sensitivity) pairs; note that recall and sensitivity are synonyms. A PR curve can be summarized by the mean Average Precision (mAP), defined as the area under the PR curve. Two types of pixel-level evaluations were performed:
\begin{description}[noitemsep]
  \item[lesion-specific\added{:}] each foreground probability map $P_m$, $m \in \{ 2, ..., M \}$, is compared to each lesion-specific segmentation map. The label $m \in \{ 2, ..., M \}$ maximizing the AUC is retained. We measure how well each lesion type can be detected by a foreground probability map.
  \item[combined\added{:}] the combined foreground probability map ($1 - P_1$) is compared to the union of all lesion segmentation maps. Here, we evaluate foreground/background separation.
\end{description}

\subsubsection{Qualitative Pixel-Level Evaluation}
\label{sec:ColorSummarizationPixelLabeling}

For improved visualization, pixel labeling can be summarized by color codes. In \explain, each pixel $I_{x,y}$ is labeled by an $M$-dimensional vector: $P_{m,x,y}$, $m \in \{ 1, ..., M \}$: this vector can be summarized by a 3-dimensional color vector in CIE L*a*b* color space. The lightness component L* represents foreground probability $1-P_{1,x,y}$. Chromaticity components a* and b* represent the normalized $(N-1)$-dimensional label vectors $P^*_{m,x,y}$:
\begin{equation}
    P^*_{m,x,y} = \left\lbrace \frac{P_{m,x,y}}{1 - P_{1,x,y}}, \forall m \in \{ 2, ..., M \} \right\rbrace \;.
\end{equation}
This normalized vector is summarized by a 2-D vector (a*, b*) using isometric mapping \citep{tenenbaum_global_2000}. Isometric mapping, or Isomap, is a manifold learning algorithm that seeks a lower-dimensional embedding maintaining geodesic distances between all input vectors.

\subsubsection{Results}

\begin{figure}[h]
  \begin{center}
    \includegraphics[width=\textwidth]{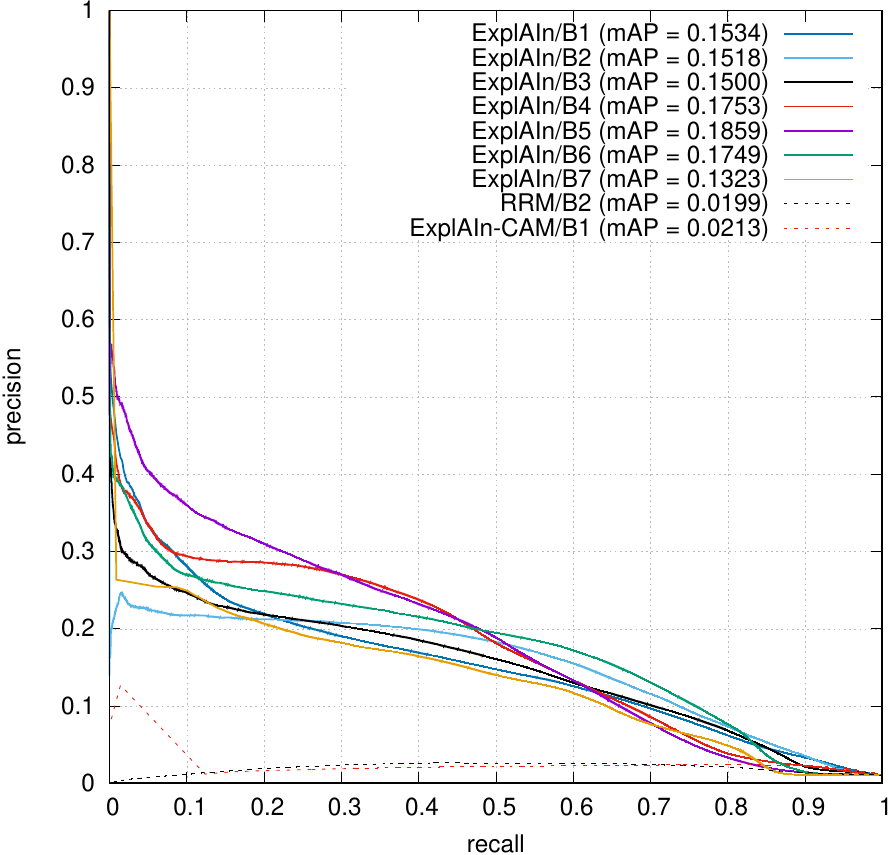}
  \end{center}
  \caption{Precision-recall analysis for DR lesion detection, i.e. pixel-level classification, in the IDRiD \citep{porwal_idrid:_2019} evaluation dataset. \added{The ``/Bx'' suffix in the architecture names indicates the EfficientNet-``Bx'' model used as backbone.}}
  \label{fig:pixelLevelPR}
\end{figure}

\begin{figure*}[!h]
  \begin{center}
    \includegraphics[width=.985\textwidth]{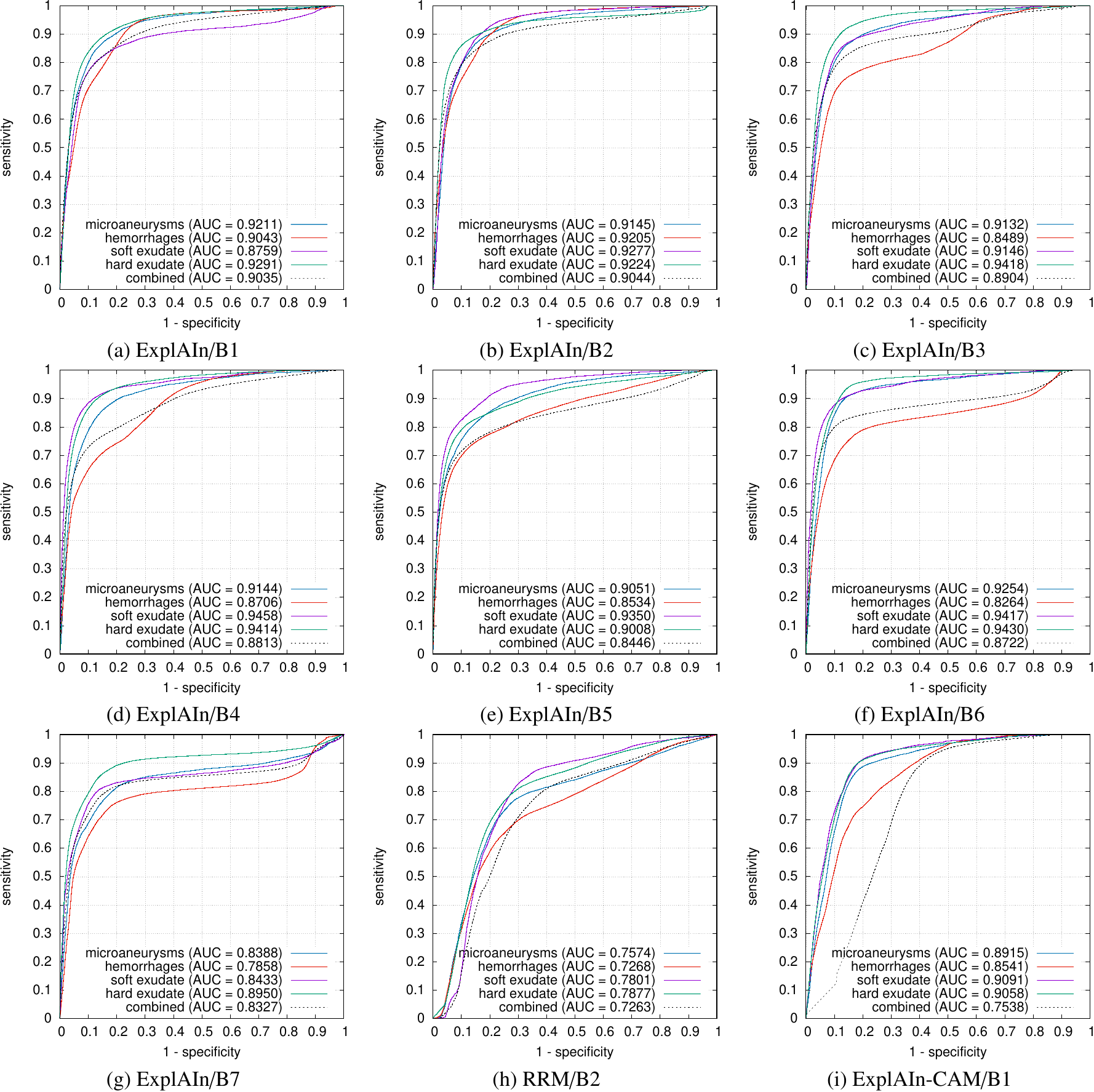}
  \end{center}
  \caption{Receiver-operating characteristics for DR lesion detection, i.e. pixel-level classification, in the IDRiD evaluation dataset. \added{The ``/Bx'' suffix in the architecture names indicates the EfficientNet-``Bx'' model used as backbone.}}
  \label{fig:pixelLevelROC}
\end{figure*}

Classification performance at the pixel level in the \changed{$IDRiD_{pixel}$ evaluation dataset} is summarized in Fig. \ref{fig:pixelLevelPR} (PR curves) and \ref{fig:pixelLevelROC} (ROC curves) for all architectures.

\begin{figure*}[h]
  \begin{center}
    \includegraphics[width=.9\textwidth]{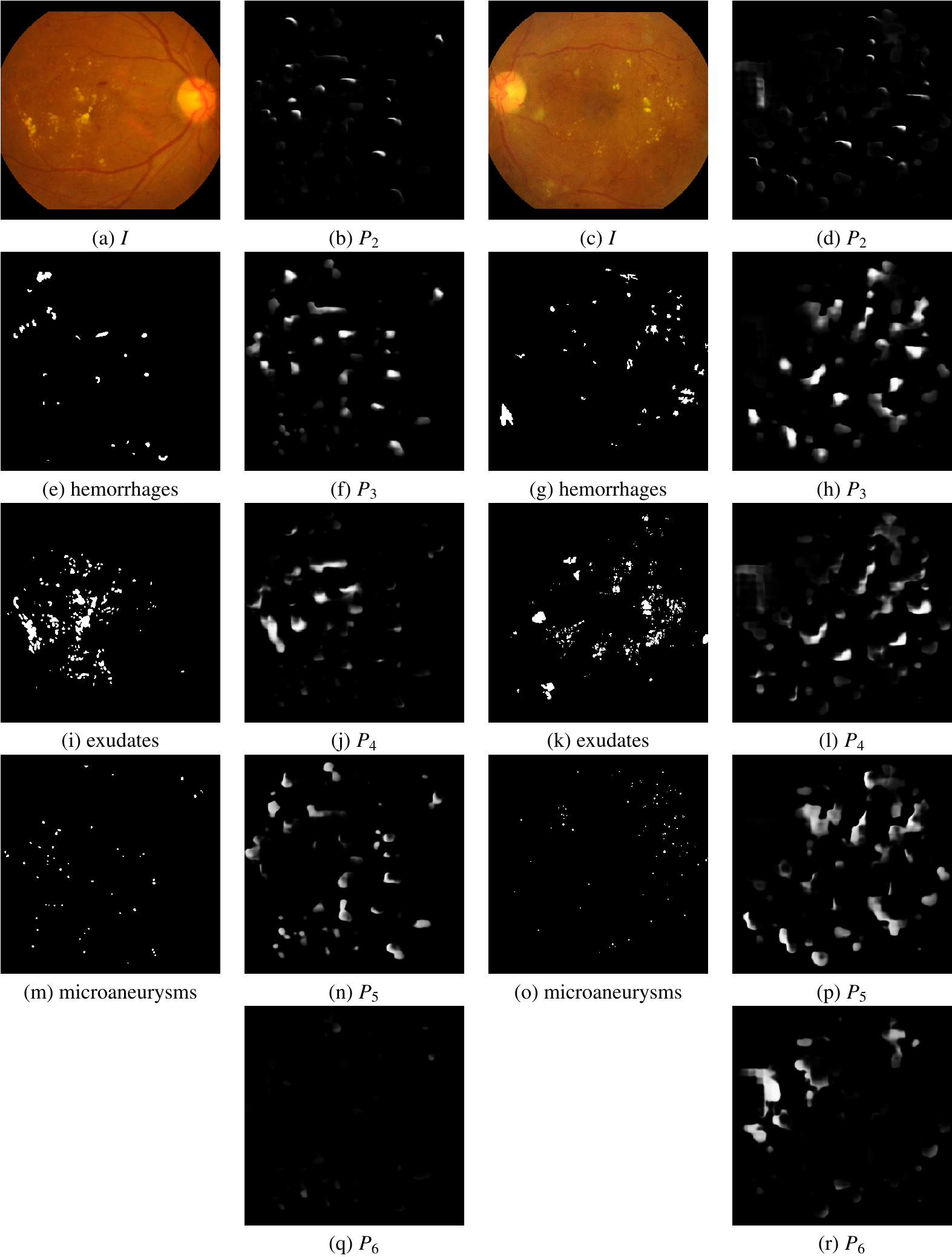}
  \end{center}
  \caption{Pixel-level predictions for images from the IDRiD evaluation dataset, using \changed{the EfficientNet-B5 backbone and $M=6$ pixel-level labels}. For each pre-processed image ($I$), manual segmentations (hemorrhages, exudates and microaneurysms) are listed below, and the $M-1$ foreground label maps ($P_2$, $P_3$, \changed{$P_4$,} $P_5$ and $P_6$) are listed on the right.}
  \label{fig:pixelLevelPredictions}
\end{figure*}

\begin{figure*}[h]
  \begin{center}
    \includegraphics[width=.875\textwidth]{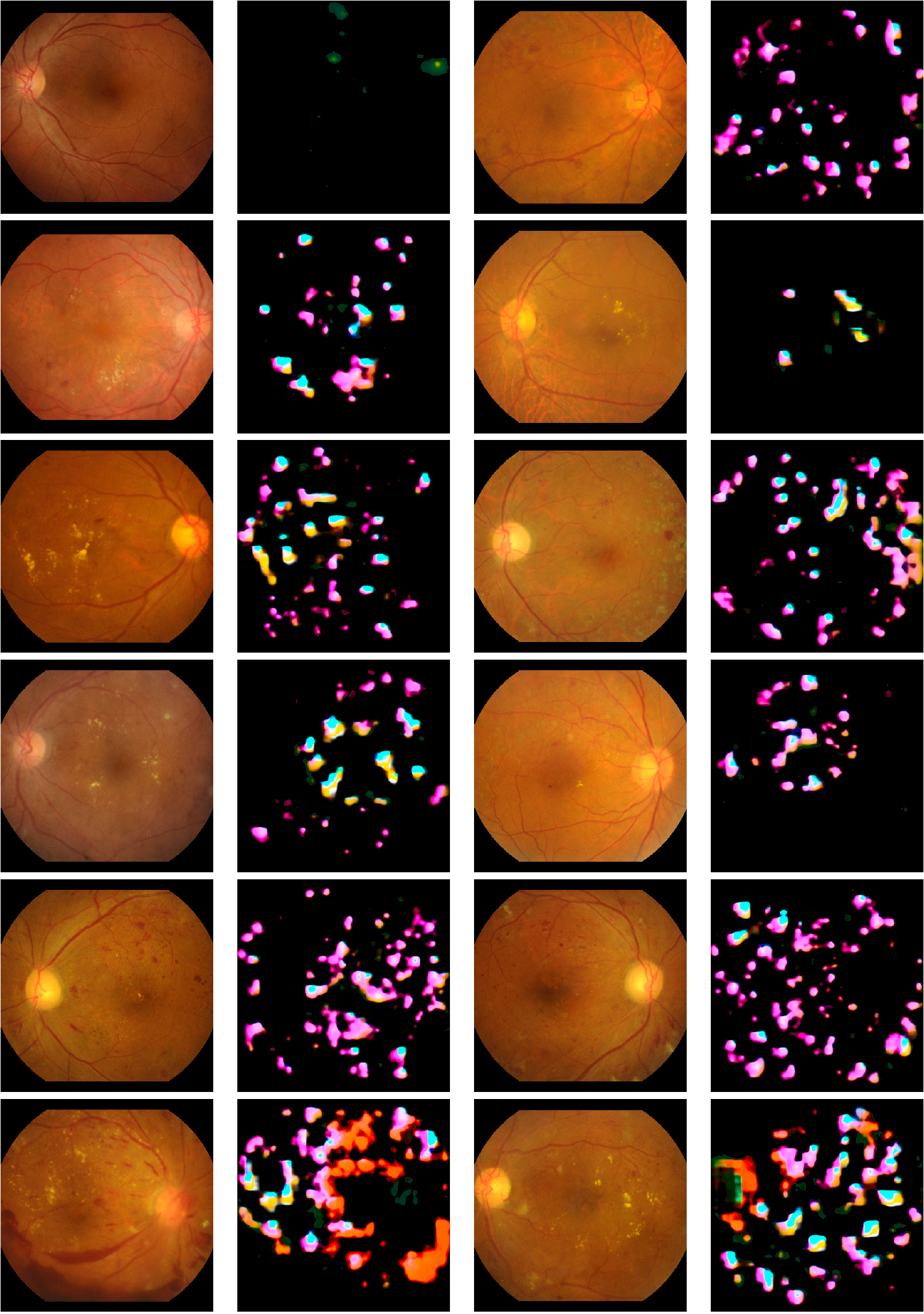}
  \end{center}
  \caption{Color-coded pixel-level predictions for images from the IDRiD evaluation dataset, using \changed{the EfficientNet-B5 backbone and $M=6$ pixel-level labels}. For each pre-processed image on the left, Color-coded pixel-level predictions (see section \ref{sec:ColorSummarizationPixelLabeling}) are given on the right.}
  \label{fig:colorPredictions}
\end{figure*}

\begin{figure*}[h]
  \begin{center}
    \includegraphics[width=.95\textwidth]{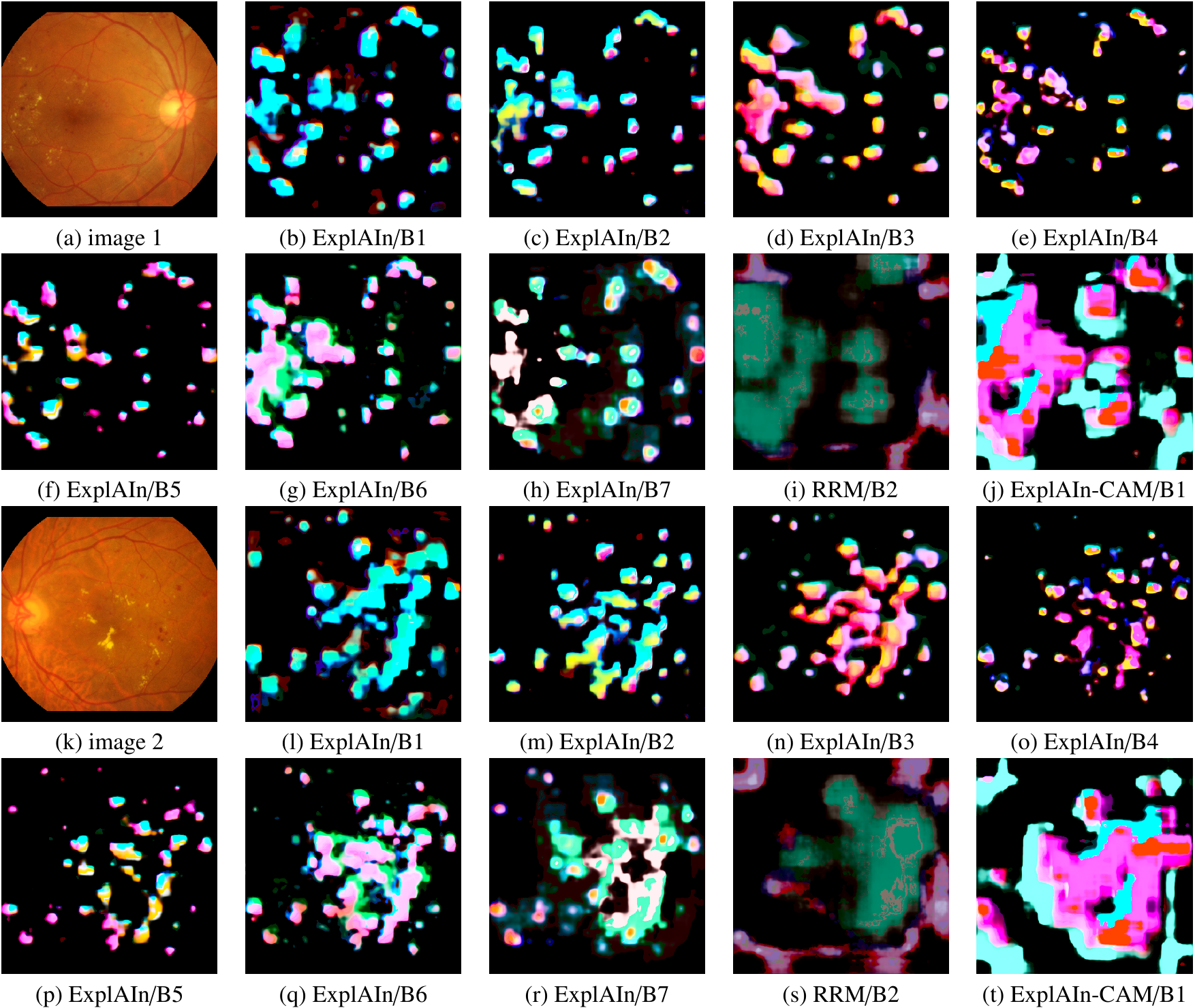}
  \end{center}
  \caption{Color-coded pixel-level predictions for two images from the IDRiD evaluation dataset using various architectures.}
  \label{fig:colorComparison}
\end{figure*}

Typical pixel-level classification maps, obtained with the optimal hyperparameters, are reported in Fig. \ref{fig:pixelLevelPredictions}. From a visual inspection, it appears that hemorrhages are targeted by map $P_3$, exudates are targeted by map $P_4$ \changed{and microaneurysms are targeted by map $P_5$.} This is confirmed by the lesion-specific quantitative evaluation described in section \ref{sec:QuantitativePixelLevelEvaluation}, the results of which are summarized in Fig. \ref{fig:pixelLevelROC} (e). \changed{Map $P_6$ seems to target signs of advanced DR, such as neovascularization.} \added{In a few images from the IDRiD dataset, we observe that $P_6$ also detects laser scars, a possible sign that neovascularization is being treated (IDRiD is not a screening dataset).} Unfortunately, advanced DR signs are not manually segmented in IDRiD, so quantitative evaluation is impossible in this case. For improved visualization, color-coded pixel-level classification maps are reported in Fig. \ref{fig:colorPredictions}. For comparison purposes, typical pixel-level classification maps obtained with different EfficientNet backbones and with the baseline methods are reported in Fig. \ref{fig:colorComparison}.

\subsection{\added{Analysis of the Extracted Rules}}

\added{Pixel-level classification maps $P_m$ have now been associated with lesion types. The next step is to verify that the relationships between these maps and image-level labeling are consistent with the ICDR scale, reported at the beginning of section \ref{sec:Experiments}. Table \ref{tab:rules} summarizes these relationships: it reports the $w_{m,n}^2$ coefficients, which are multiplied by the maximal $P_{m,x,y}$ values, in order to obtain the image-level probabilities $p_n$ in Eq. (\ref{eq:deltaLayer}). We see that the dominant map for detecting `at least mild NPDR' is $P_5$, which we associated with microaneurysms. The dominant map for detecting `at least moderate NPDR' is $P_4$, which we associated with exudates. The dominant map for detecting `at least severe NPDR' is $P_3$, which we associated with hemorrhages. Finally, the dominant map for detecting `PDR' is $P_6$, which we associated with advanced DR signs, including neovascularization in particular. Overall, these relationships are consistent with the ICDR scale: the targeted lesions are compatible.}

\begin{table*}[h]
    \caption{\added{Relationship between pixel-level classifications and image-level labeling for DR grading. This table reports the $w_{m,n}^2$ coefficients of Eq. (\ref{eq:deltaLayer}), which link the maximal intensity in each map $P_m$ to the image-level probability $p_n$, as well as the bias term $b_n$. For each rule, the dominant coefficient is in bold.}}
    \added{
    \begin{center}
      \begin{tabular}{c|ccccc|c}
                                         & \multicolumn{5}{c|}{$w_{m,n}^2$ coefficients}  & $b_n$ bias \\
          \hline
          $p_n \backslash P_m$           & $P_2$ & $P_3$ & $P_4$ & $P_5$ & $P_6$          &            \\
          \hline
          at least mild NPDR ($p_1$)     & 3.009 & 1.673 & 1.070 & \textbf{3.892} & 0.417 & -2.765 \\
          at least moderate NPDR ($p_2$) & 1.288 & 2.024 & \textbf{3.159} & 2.417 & 0.915 & -4.024 \\
          at least severe NPDR ($p_3$)   & 0.257 & \textbf{4.294} & 0.137 & 2.248 & 1.268 & -5.318 \\
          PDR ($p_4$)                    & 0.000 & 2.141 & 0.000 & 0.592 & \textbf{4.235} & -5.885 \\
      \end{tabular}
    \end{center}
    }
    \label{tab:rules}
\end{table*}

\added{Map $P_6$ could not be validated at the pixel level, due to lack of manual annotations. However, Table \ref{tab:rules} shows that $P_6$ is predominantly linked to PDR detection. In details, PDR detection relies on $P_6$ and, to a lesser extent, $P_3$. One possible interpretation is that $P_3$, which was shown to detect hemorrhages, detects `vitreous/preretinal hemorrhages' and that, therefore, $P_6$ detects the other sign of PDR, namely neovascularization \citep{wilkinson_proposed_2003}.}

\section{Discussion and Conclusions}
\label{sec:DiscussionConclusions}

We have presented \explain, a novel eXplanatory Artificial Intelligence (XAI) framework for multilabel image classification. In addition to labeling images, \explain~also classifies each pixel within images. Simple rules link pixel-level classification to image-level labeling. Consequently, image-level labeling can be explained simply by pixel-level classification. Unlike image-level labeling, pixel-level classification is self-supervised; a novel occlusion method is presented to ensure satisfactory foreground/background pixel separation and therefore meaningful explanations. This framework was applied to the diagnosis of Diabetic Retinopathy (DR) using Color Fundus Photography (CFP). Classification performance was evaluated both at the image level and at the pixel level.

\explain~models were trained to classify DR at the image level, using data from French and American DR screening programs (OPHDIAT and EyePACS): the goal was to grade DR severity in one eye according to the \changed{ICDR} scale. On an Indian dataset (IDRiD), we found that manually-segmented DR lesions (microaneurysms, hemorrhages, soft exudates and hard exudates) could be detected well (see Fig. \ref{fig:pixelLevelROC}). Overall, these lesions could also be differentiated correctly (see Fig. \ref{fig:pixelLevelPredictions}). Using the optimal number of pixel-level labels ($M=6$) and the optimal CNN backbone (EfficientNet-B5), unsupervised pixel-level classification can be schematized as follows. By design, label $m=1$ was assigned to the background (see section \ref{sec:LearningDetectBackgroundPixels}). One pixel-level label was assigned to microaneurysms (\changed{$m=5$}), another one was assigned to hemorrhages ($m=3$), another one to exudates ($m=4$) and another one to advanced DR signs (\changed{$m=6$}). The two types of exudates (hard and soft), however, were not separated well using this optimal CNN backbone. Assigning pixel-level labels to lesion types makes sense, since the \changed{ICDR} scale relies primarily on the types of DR lesions present in images \citep{wilkinson_proposed_2003}. Regarding the last label, $m=2$, it seems to group together false alarms of the occlusion method: pixels irrelevant for DR classification but nevertheless unusual for background pixels. \added{Table \ref{tab:rules} suggests that the learnt relationships between pixel-level classification and image-level labeling is consistent with the ICDR scale.} For easier visualization, we propose to analyze all pixel-level labels jointly, using color codes obtained through dimension reduction (see section \ref{sec:ColorSummarizationPixelLabeling}). One advantage is a more compact representation: one color-coded image is generated per input image, as opposed of $M=6$ grayscale images. \changed{As illustrated in Fig. \ref{fig:colorComparison}, where different CNN backbones are compared, it appears that changing model hyperparameters does not change image-level classification radically, although large CNN backbones (EfficientNet models B4 and higher) lead to semantically richer} pixel-level classification: color codes are more diverse.

\added{Pixel-level ROC curves in Fig. \ref{fig:pixelLevelROC} indicate that lesions can be detected with a good pixel-level sensitivity: under-segmentation is limited. On the other hand, pixel-level PR curves in Fig. \ref{fig:pixelLevelPR} indicate that pixel-level precision is lower: over-segmentation is more problematic. However, we believe this level of precision is enough for explainability purposes.} \changed{In any case, \explain~clearly is superior to \explain-CAM and RRM at the pixel level} (see Fig. \ref{fig:colorComparison}). In particular, precision is about 10 times higher (see Fig. \ref{fig:pixelLevelPR}). In \explain, increasing $\gamma$, the weight assigned to the $\mathcal{L}_{sparsity}$ loss in Eq. (\ref{eq:totalLoss}), increases pixel-level precision without dramatically impacting image-level classification. On the other hand, enforcing the sparsity of CAM, like in \explain-CAM, largely impacts image-level classification. This is due to the low resolution of CAM: a sparse CAM implies that large regions of images have to be completely ignored. This suggests that CAM is not precise enough for Weakly-Supervised Semantic Segmentation (WSSS) of DR-related lesions, even though this is currently the most frequently used technique for WSSS (see section \ref{sec:RelatedMachineLearningFrameworks}).

\added{We have shown that \explain~detects and categorizes relevant abnormalities for grading DR severity. But the type of abnormalities is not the only relevant factor. In the ICDR scale, for instance, severe NPDR is only triggered when abnormalities are sufficiently numerous and sufficiently spread across the retina. How can \explain~reproduce these requirements, given that it only looks at the maximal intensity in each pixel-level classification map?} \changed{Different solutions were witnessed during the course of training. One solution was to assign one pixel-level label per quadrant, in order to capture spatial distribution. Another solution was to focus on large clusters of abnormalities and ignore isolated abnormalities, in an attempt to capture their number. The latter solution was retained in the \explain-CAM baseline, as illustrated in Fig. \ref{fig:colorComparison} (j) and (t). In \explain, however, none of these solutions was retained at the end of training. Each type of abnormalities is assigned to a single pixel-level label and most of the abnormalities are detected (see Fig. \ref{fig:pixelLevelPredictions}): pixel-level probabilities simply tend to increase with the number of abnormalities and with their spread across the image.}

Although explainability is a desirable feature, it should not come at the expense of decreased classification performance. Fortunately, for the same task on the same evaluation dataset \added{($OPHDIAT_{eval}$)}, image-level classification performance of \explain~(see Fig. \ref{fig:imageLevelROC}) is similar to previously reported results for a ``black-box'' AI solution from our group \citep{quellec_instant_2019}. Clearly, this ``black-box'' solution was designed without explainability constraints: in particular, it consisted of an ensemble of multiple CNNs while, for explainability purposes, \explain~necessarily consists of a single CNN. In details, two classification criteria have improved in \explain~(see Table \ref{tab:imageLevelAUC}):
\begin{itemize}[noitemsep]
  \item ``\changed{at least moderate NPDR}'', the most important criterion in screening applications: AUC increases from 0.9859 to 0.9939. Besides, for a 100\% sensitivity, specificity increases from 12.5\% to 88.1\%,
  \item ``\changed{at least severe NPDR}'': AUC increases from 0.9969 to 0.9978. Besides, for a 100\% sensitivity, specificity increases from 92.4\% to 95.6\%.
\end{itemize}
For the remaining two criteria, performance in terms of AUC decreases overall. However, the optimal sensitivity/specificity endpoint is improved or unchanged:
\begin{itemize}[noitemsep]
  \item ``\changed{at least mild NPDR}'': sensitivity = 90\%, specificity = 93.3\% (``black-box'' AI) or 94.1\% (\explain),
  \item ``PDR'': sensitivity = 100\%, specificity = 98.9\% (in both solutions).
\end{itemize}
We note that \explain~is also more efficient than \explain-CAM for classification at the image level (see Table \ref{tab:imageLevelAUC}), although the difference is more subtle than at the pixel level. \added{\explain~was also evaluated on two independent datasets: the Indian IRDiD dataset and the Chinese DeepDR dataset. The proposed algorithm seems to compare favorably with baseline algorithms, even though these baseline algorithms were trained or fine-tuned on non-independent training datasets (see Table \ref{tab:imageLevelAccKappa}). Note that the comparison is not direct on DeepDR: the proposed and baseline algorithms were evaluated on slightly different datasets (validation and onsite evaluation datasets). As shown in Table \ref{tab:imageLevelAccKappa}, the accuracy of \explain~varies across evaluation datasets. These variations can be explained by several factors: a different distribution of DR severity levels, different annotators (the OPHDIAT evaluation images were read by at least two graders), different population characteristics, different acquisition devices, etc. Variations in terms of quadratic weighted Kappa ($\kappa$) are more limited. Importantly, a large $\kappa$ value was measured in all evaluation datasets considered in this paper ($0.8673 \leq \kappa \leq 0.9243$), indicating that disagreements between the AI and experts generally have a limited amplitude.}

The very good classification performance of \explain~at the image level was somehow unexpected. One possible explanation for this success is that the proposed generalized occlusion method acts as a regularization operator. Enforcing foreground sparsity encourages the AI to look for local patterns in images rather than to analyze the image globally. More generally, by solving a multi-task problem (image-level and pixel-level classification), we favor the extraction of more general and relevant features.

In this paper, \explain~has been applied to a well-known classification problem (DR severity classification). It allowed us to evaluate the relevance of the identified local patterns (namely DR lesions). However, \explain~would be even more useful for a totally new classification problem (disease progression prediction, diagnosis of a new disease, etc.): it would allow knowledge acquisition. In medicine, for instance, it would help clinicians quickly identify new useful markers in images.

One limitation of the proposed solution for foreground/background separation (the generalized occlusion method) is that it is limited to binary or multilabel classification (zero, one or multiple labels per image): it is not directly applicable to multiclass classification (exactly one label per image). This is because it relies on the concept of ``background image'', which does not generally apply to multiclass classification. \added{This concept may apply to a subclass of multiclass classification problems where one of the classes corresponds to background images (with no relevant patterns visible); however, the occlusion loss would need a different formulation.} \changed{Finally, the validation of \explain~on the DR grading task also has one limitation:} pixel-level evaluation relies on incomplete lesion segmentations. Only four types of lesions were manually-segmented in IDRiD: microaneurysms, hemorrhages, soft exudates and hard exudates \citep{porwal_idrid:_2019}. This results in pessimistic precisions in Fig. \ref{fig:pixelLevelPR}.

In conclusion, we have presented a novel explanatory AI framework for multilabel image classification. Given an image, this framework 1) localizes and categorizes relevant patterns in the image, 2) classifies the image and 3) explains how exactly each pattern contributes to the classification. If experts can assign keywords to these patterns, then a textual report can also be generated. For the task of DR diagnosis using CFP, the new explainability feature comes without loss of classification performance. Thanks to this new feature, we expect healthcare AI systems to gain the trust of clinicians and patients more easily, which would facilitate their deployment.

\bibliographystyle{elsarticle-harv}
\bibliography{ExplAIn}

\end{document}